\documentclass[preprint,12pt]{elsarticle}

\usepackage{amsmath,amssymb}

\usepackage{graphicx}
\usepackage{booktabs}
\usepackage{array}
\usepackage{multirow}
\usepackage{float}

\usepackage{tikz}
\usetikzlibrary{shapes.geometric, arrows.meta, positioning, fit, backgrounds}

\usepackage{url}
\usepackage{xcolor}
\usepackage{hyperref}

\hypersetup{
  colorlinks=true,
  linkcolor=blue,
  citecolor=blue,
  urlcolor=blue
}

\journal{Information Sciences}

\begin{document}

\begin{frontmatter}



\title{Graph Neural Networks for Scalable and Transferable Node Centrality Approximation}

\author[insubria,uhasselt]{Samra Sana\corref{cor1}}
\ead{samra.sana@studenti.uninsubria.it}

\author[insubria]{Giorgio Mantica}
\author[insubria]{Saul Imbrici}

\cortext[cor1]{Corresponding author}

\affiliation[insubria]{
organization={Center for Nonlinear and Complex Systems, Universit\`a degli Studi dell'Insubria},
city={Como},
country={Italy}
}

\affiliation[uhasselt]{
organization={Data Science Institute, Hasselt University},
city={Hasselt},
country={Belgium}
}

\begin{abstract}
Graph Neural Networks (GNNs) provide a learning-based framework for approximating graph quantities that are expensive to compute exactly. This paper investigates GNNs for scalable approximation of betweenness and closeness centrality, formulated as a node-ranking problem. Exact centrality values are used as supervision, and ranking quality is evaluated using Kendall's $\tau$ rank correlation. We study whether message-passing GNNs can learn transferable structural representations across different graph topologies rather than only fitting the distribution used during training. On unseen Erd\H{o}s--R\'enyi graphs, the proposed models achieve $\tau = 0.851$ for betweenness and $\tau = 0.894$ for closeness. A large-scale betweenness model trained on graphs with $N = 5{,}000$ nodes achieves $\tau = 0.938$, demonstrating scalability. Mixed-distribution training on Erd\H{o}s--R\'enyi, Barab\'asi--Albert, and Gaussian Random Partition graphs improves betweenness transfer across graph families. In contrast, closeness centrality remains more sensitive to community-structured graphs and shows reduced transfer to real-world topologies. Finally, GNN inference achieves up to a $97.7\times$ speedup over exact computation. These results show that mixed-distribution training can improve structural transfer in GNN-based centrality approximation, while identifying closeness centrality's sensitivity to topology as an open challenge.

\end{abstract}


\end{frontmatter}



\section{Introduction}

Graph Neural Networks (GNNs) have become an important class of learning systems for graph-structured data. By combining local message passing with trainable nonlinear transformations, GNNs can learn node representations that support tasks such as node classification, link prediction, graph classification, and ranking. Beyond standard prediction tasks, GNNs are increasingly used as neural approximators for graph algorithms, where the objective is to learn graph quantities that are expensive to compute exactly.
Node centrality approximation is a natural and challenging problem in this setting. Centrality measures quantify the structural importance of nodes in a graph and are widely used in social, biological, transportation, communication, and recommendation networks~\cite{newman2010}. Among them, betweenness centrality measures how often a node lies on shortest paths between other nodes and is associated with brokerage or control of information flow~\cite{freeman1977}. Closeness centrality measures how efficiently a node can reach the rest of the network~\cite{sabidussi1966,newman2010}. However, exact computation of these measures becomes expensive for large graphs, especially for betweenness centrality, which requires repeated shortest-path computations.

\begin{figure}[H]
\centering
\includegraphics[width=0.80\textwidth]{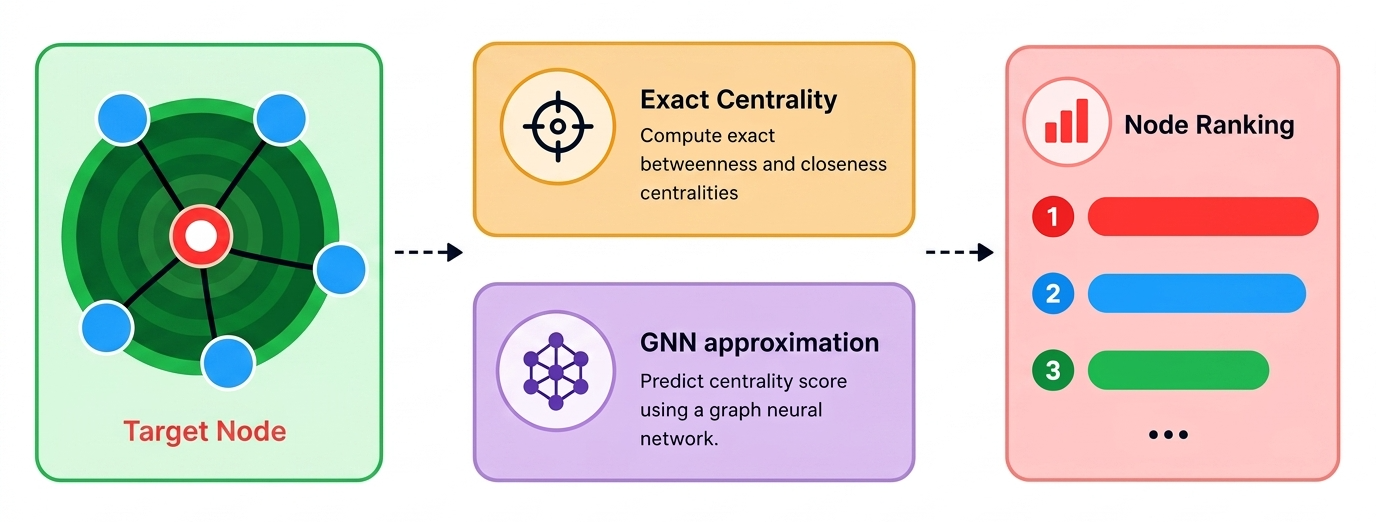}
\caption{\textbf{Goal:} approximate the ranking induced by exact betweenness and closeness centrality.}
\label{fig:goal}
\end{figure}

This computational cost motivates learning-based approximation methods that can preserve centrality-induced node rankings while reducing inference time. In this work, we formulate betweenness and closeness centrality approximation as supervised node-ranking problems. Instead of reproducing exact centrality values, the goal is to learn the ranking induced by exact centrality measures, as illustrated in Figure~\ref{fig:goal}. This formulation is suitable for many applications in which the relative importance of nodes is more relevant than the exact numerical centrality score.

Message-passing GNNs provide a data-driven framework for learning such rankings directly from graph topology~\cite{hamilton2017,kipf2017}. Previous studies have shown that GNNs can approximate centrality measures efficiently~\cite{maurya2021}. However, an important question remains insufficiently understood: do GNNs learn transferable structural representations for centrality approximation, or do they mainly exploit properties of the graph distribution used during training?

This question is particularly important because real-world networks often differ substantially from the synthetic graph families used for training. A model trained on homogeneous random graphs may not generalize to scale-free, community-structured, spatial, or biological networks. 

This paper investigates the scalability and transferability of GNN-based centrality approximation across synthetic graph families and real-world networks. We evaluate whether models trained on one graph distribution can generalize to structurally different networks, and whether mixed-distribution training improves robustness. The study considers Erd\H{o}s--R\'enyi, Barab\'asi--Albert, and Gaussian Random Partition graphs, together with zero-shot evaluation on real-world networks including the \textit{C. Elegans} neural network, Email-Eu-Core, and the Western US power grid.

The main contributions of this paper are as follows:
\begin{enumerate}
\item We formulate betweenness and closeness centrality approximation as supervised node-ranking problems using Graph Neural Networks.

\item We design and evaluate message-passing GNN models for learning centrality-based node rankings directly from graph topology.

\item We study cross-topology generalization across Erd\H{o}s--R\'enyi, Barab\'asi--Albert, and Gaussian Random Partition graphs.

\item We show that mixed-distribution training improves betweenness transfer, suggesting that structural diversity supports more robust graph representation learning.

\item We evaluate zero-shot transfer on real-world networks, including the \textit{C. Elegans} neural network, Email-Eu-Core, and the Western US power grid.

\item We quantify the computational advantage of GNN-based approximation, showing up to a $97.7\times$ inference speedup over exact betweenness computation.

\end{enumerate}

\section{Background and Related Work}
\label{sec:background}

\subsection{Network Centrality Measures}

A network is represented as a graph $G = (\mathcal{N}, \mathcal{E})$, where $\mathcal{N}$ is the set of nodes and $\mathcal{E}$ is the set of edges connecting them. Centrality assigns a score to each node that reflects its structural importance. However, importance may correspond to different roles depending on the network and the application.

\textbf{Betweenness centrality} measures how often a node lies on shortest paths between other nodes. For node $i$, it is defined as
\begin{equation}
b_i = \sum_{j \neq k} \frac{\sigma_{jk}(i)}{\sigma_{jk}},
\label{eq:betweenness}
\end{equation}
where $\sigma_{jk}$ is the total number of shortest paths between nodes $j$ and $k$, and $\sigma_{jk}(i)$ counts only those passing through node $i$~\cite{freeman1977}. A node with high betweenness acts as a broker; removing it can make many shortest paths longer or disconnect them entirely.

\textbf{Closeness centrality} measures how efficiently a node can reach the rest of the network:
\begin{equation}
c_i = \frac{N-1}{\sum_j d_{ij}^{\min}},
\label{eq:closeness}
\end{equation}
where $d_{ij}^{\min}$ is the shortest-path distance from node $i$ to node $j$~\cite{newman2010,sabidussi1966}. A node with high closeness is located near the structural center of the network.

Both measures depend on shortest-path computations across the graph and become computationally demanding as network size increases. Although efficient exact algorithms exist, their runtime remains a bottleneck for large-scale networks in biology, infrastructure, communication, and social systems. This motivates approximation methods that preserve centrality-induced rankings while reducing computational cost. In this paper, centrality approximation is treated as a learning problem in which a neural model predicts the relative importance of nodes directly from graph topology.

\subsection{Graph Neural Networks}
Graph Neural Networks extend neural-network learning to graph-structured data~\cite{hamilton2017,kipf2017}. Unlike multilayer perceptrons, which process input features independently, GNNs explicitly model relationships between nodes through message-passing operations:
\begin{align}
a_v^{(k)} &= \mathrm{AGGREGATE}^{(k)}!
\left({h_u^{(k-1)} : u \in \mathcal{N}(v)}\right), \\
h_v^{(k)} &= \mathrm{COMBINE}^{(k)}!
\left(h_v^{(k-1)},, a_v^{(k)}\right),
\end{align}
where $h_v^{(k)}$ is the representation of node $v$ at layer $k$, and $\mathcal{N}(v)$ denotes the neighbours of $v$. After $L$ layers, the representation $h_v^{(L)}$ contains information from the $L$-hop neighbourhood of node $v$. Thus, increasing the number of layers enlarges the receptive field, allowing the model to encode broader structural information.

Modern GNNs are commonly formulated within the Message Passing Neural Network framework, where node representations are iteratively updated through neighbourhood aggregation. Prominent examples include Graph Convolutional Networks, which use spectral-inspired aggregation operators, and GraphSAGE, which supports inductive learning through neighbourhood sampling. These architectures have shown strong performance in node classification, link prediction, graph representation learning, and ranking tasks. More recently, they have also been studied as neural approximators for graph algorithms, where the objective is to learn computationally expensive graph quantities from examples. This makes GNNs a suitable framework for approximating centrality-based node rankings.

\subsection{Related Work}

Centrality approximation has traditionally been addressed using exact graph algorithms and sampling-based methods. Brandes algorithm remains the standard exact method for betweenness centrality~\cite{brandes2001}, while later approximation approaches reduce computational cost by sampling source nodes, shortest paths, or graph substructures~\cite{bader2007,riondato2014,borassi2016}. These methods provide theoretical or empirical efficiency gains, but their accuracy depends on the sampling budget and on the structural properties of the graph.

A second line of work studies neural and graph-learning approaches for graph-theoretic and algorithmic problems. GNNs provide a natural framework for learning structural representations through message passing~\cite{kipf2017,hamilton2017,battaglia2018}. Prior work has shown that neural models can learn to identify highly central nodes~\cite{fan2019}, and Maurya et al.~\cite{maurya2021} proposed GNN models for fast approximation of betweenness and closeness centrality. More broadly, neural algorithmic reasoning investigates whether neural networks can learn algorithmic procedures such as shortest paths, sorting, and dynamic programming~\cite{velickovic2022}. This perspective is closely related to centrality approximation, since both betweenness and closeness depend on shortest-path structure.

Recent work has further extended learning-based centrality approximation. Zhang et al.~\cite{zhang2022nodeimportance} proposed a deep-learning approach for evaluating node importance in complex networks, confirming the relevance of neural models for node-importance estimation. Zou et al.~\cite{zou2024cnca} proposed CNCA-IGE, an inductive encoder--decoder model for approximating closeness and betweenness centrality rankings using GraphSAGE, a variational graph autoencoder, and an MLP-Mixer decoder. Their approach emphasizes inductive representation learning across synthetic and real-world networks. Dachille et al.~\cite{dachille2026brava} introduced BRAVA-GNN, a recent betweenness-ranking model based on multi-hop degree-mass features and synthetic training graphs designed to improve generalization to high-diameter real-world networks.

A third relevant direction concerns transferability and distribution shift in graph learning. GNNs are often trained and tested on graphs drawn from similar distributions, but their performance can degrade when graph topology changes~\cite{yehuda2020,wu2022ood,gui2022good}. This issue is particularly important for centrality approximation, because the structural meaning of node importance depends on global network topology. Recent centrality-approximation studies have addressed inductive learning and generalization, but the role of training-distribution diversity itself remains less explicitly isolated.

In contrast to prior work, the present study focuses on training-distribution diversity as a factor governing transfer. We compare single-topology training with mixed-topology training across Erd\H{o}s--R\'enyi, Barab\'asi--Albert, and Gaussian Random Partition graphs, and show a quantified improvement in betweenness transfer on community-structured graphs. We also evaluate both betweenness and closeness centrality, showing that closeness is substantially more sensitive to topology shift. This indicates that transfer behaviour depends not only on the GNN architecture, but also on the centrality measure and the structural diversity of the training distribution.

\subsection{Approximate Centrality Baselines}

The most widely used exact framework for betweenness computation is Brandes' algorithm~\cite{brandes2001}. A common approximation strategy samples $k$ pivot nodes and builds shortest-path trees only from those pivots, reducing the cost to approximately $O(k \cdot N \cdot E)$. At small $k$, this approach is fast but less accurate; at larger $k$, its accuracy improves but its cost approaches that of exact computation. The performance of sampling-based methods therefore depends strongly on the number and selection of sampled nodes. In practice, choosing an appropriate sampling budget requires balancing computational cost against ranking accuracy, and the optimal choice may vary substantially across network types.

A much simpler baseline is degree centrality, which counts the number of edges incident to each node and can be computed very efficiently. On Erd\H{o}s--R\'enyi graphs, degree centrality can be surprisingly competitive because high-degree nodes often lie on many shortest paths and tend to be close to other nodes. We include degree centrality as a baseline because it is computationally cheap and provides a useful reference point for learned approximation methods. We compare both sampling-based Brandes approximation and degree centrality against the proposed GNN models in Section~\ref{sec:baselines}.

\section{GNN Architecture and Training}
\label{sec:methodology}
This section presents the proposed GNN-based learning framework for centrality approximation, including dataset construction, model architecture, training procedure, and ranking-based supervision.

\subsection{Dataset Construction}
\label{sec:dataset}

All graphs are generated synthetically using NetworkX~\cite{hagberg2008networkx}. We consider three graph families that represent distinct structural regimes:
homogeneous random connectivity, hub-dominated connectivity, and modular community structure.

\textbf{Erd\H{o}s--R\'enyi (ER)} graphs are generated with $N$ nodes and edge probability $p$, where each possible edge is included independently with probability $p$~\cite{erdos1959}. ER graphs serve as the primary training distribution. In the main experiments, we use $N=200$ and $p=0.15$, giving an expected average degree of $29.9$. ER graphs are generated as directed for
betweenness centrality and undirected for closeness centrality.

\textbf{Barab\'asi--Albert (BA)} graphs are generated by preferential attachment with $m=\max(1,\lfloor pN/2 \rfloor)$ edges added per new node~\cite{barabasi1999}.
This produces scale-free networks with heterogeneous degree distributions and dominant hub nodes.

\textbf{Gaussian Random Partition (GRP)} graphs are generated with mean community size $s=20$, variance $v=5$, intra-community edge probability $p_{\text{in}}=0.3$, and inter-community edge probability $p_{\text{out}}=0.05$. These graphs represent modular networks with explicit community structure.

These synthetic graph families allow controlled evaluation of how GNN-learned node representations transfer across structurally different network topologies.

For the mixed betweenness experiment, we use a balanced dataset containing 533 ER graphs, 533 BA graphs, and 533 GRP graphs, for a total of 1,599 graphs.
\subsubsection{Ground-Truth Labels}
For each graph, exact betweenness and closeness centrality values are computed using NetworkX and used as supervision for learning centrality-induced node rankings, with \texttt{betweenness\_centrality(normalized=True)} and
\texttt{closeness\_centrality}. All labels for the primary $N=200$ experiments are exact. A separate scalability experiment is also performed on graphs with $N=5{,}000$ nodes.

\subsubsection{Dataset Sizes and Splits}

All datasets are divided into training, validation, and test sets using a 70\%--10\%--20\% split. The validation set is used for hyperparameter selection, while the test set is used only for final evaluation.
\small{
\begin{table}[H]
\centering
\begin{tabular}{llcccc}
\toprule
\textbf{Experiment} & \textbf{Graph types} &
\textbf{Total} & \textbf{Train} &
\textbf{Val} & \textbf{Test} \\
\midrule
Betweenness (primary)  & ER only       & 2000 & 1400 & 200 & 400 \\
Closeness (primary)    & ER only       & 5000 & 3500 & 500 & 1000 \\
Betweenness (mixed)    & ER + BA + GRP (533 each) & 1599 & 1120 & 159 & 320 \\
Betweenness ($N=5000$) & ER only       & 300  & 210  & 30  & 60 \\
\bottomrule
\end{tabular}
\caption{Dataset summary for all experiments. Primary experiments use $N=200$ and $p=0.15$.}
\label{tab:datasets}
\end{table}
}
The primary experiments use $N=200$ because exact centrality labels can be computed efficiently for many graphs, enabling controlled training and evaluation. To assess scalability beyond this setting, we additionally train and evaluate a large-scale betweenness model on graphs with $N=5{,}000$ nodes.

\subsection{Model Architecture}
\label{sec:architecture}
The proposed models treat centrality approximation as a supervised graph representation learning problem for node ranking. For betweenness centrality, we use a dual-pathway GNN architecture that learns complementary representations from outgoing and incoming graph structure, since shortest-path flow depends on directional connectivity. For closeness centrality, a single pathway is used because the model is trained on undirected graphs.

The number of message-passing layers controls the receptive field of the learned node representations. In the primary ER setting with $N=200$ and $p=0.15$, shortest-path distances are typically small, and the graph diameter is expected to be only a few hops. Therefore, $L=5$ provides sufficient neighborhood coverage for betweenness centrality while limiting oversmoothing. For closeness centrality, we use a slightly deeper model with $L=7$, since closeness depends on aggregate distances from each node to the rest of the graph and can benefit from broader neighborhood aggregation. The ablation results in Section~\ref{sec:ablation} show that increasing depth does not monotonically improve performance, indicating that $L$ must balance receptive-field size, optimization stability, and robustness under topology shift.

\begin{figure}[H]
    \centering
    \includegraphics[width=0.95\textwidth]{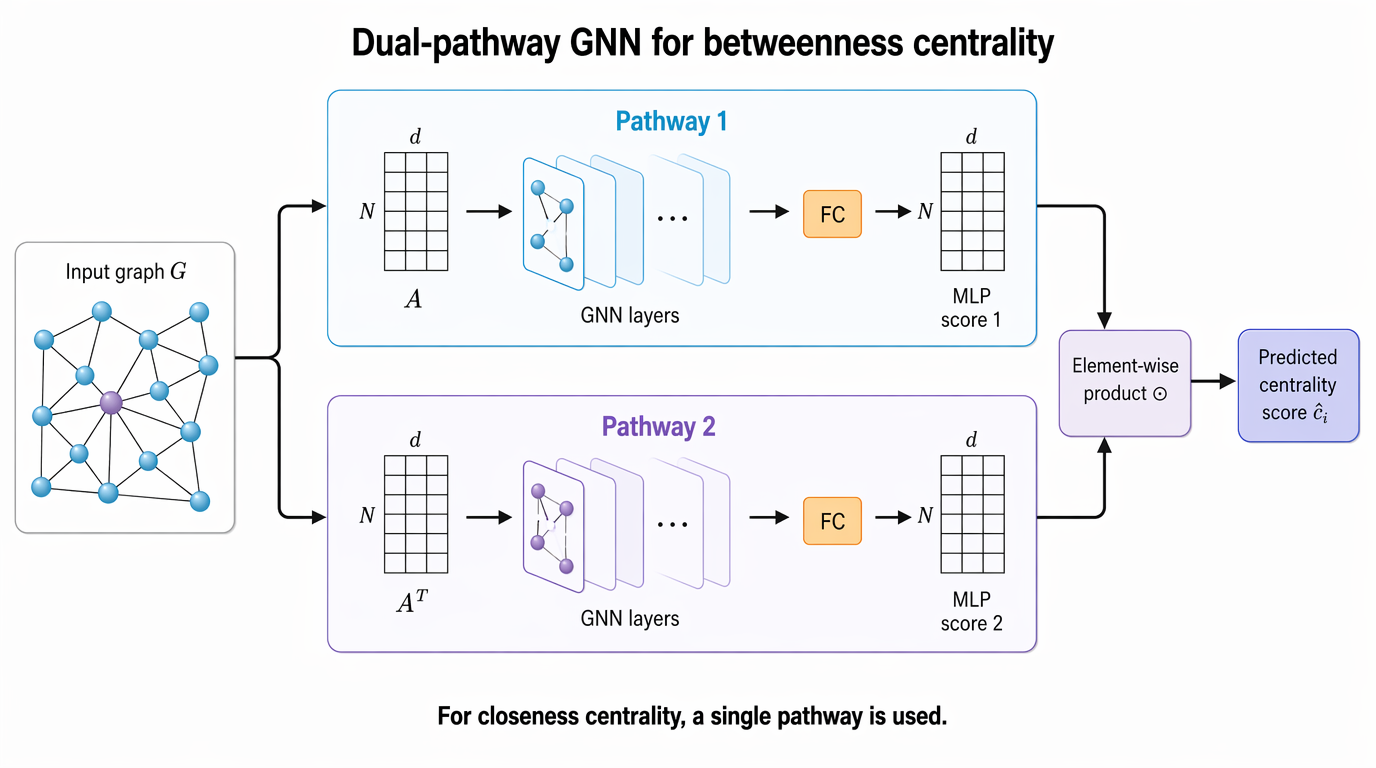}
    \caption{Dual-pathway GNN architecture for learning betweenness-based node rankings. The first pathway processes the adjacency matrix $A$, while the second pathway processes the transpose $A^{T}$. Each pathway produces an MLP score, and the final prediction is obtained by an element-wise product. For closeness centrality, only a single pathway is used.}
    \label{fig:gnn_dual_pathway}
\end{figure}

\subsubsection{Betweenness Centrality GNN}

Given a graph $G=(\mathcal{N},\mathcal{E})$ with adjacency matrix $A$, the betweenness GNN processes both $A$ and $A^T$ to learn direction-aware node representations as shown in Figure~\ref{fig:gnn_dual_pathway}. The two pathways capture complementary information about outgoing and incoming geodesic flow.
The model uses $L=5$ message-passing layers. At layer $k$, node representations
are updated as
\begin{equation}
  h_v^{(k)} =
  \mathrm{normalize}\!\left[
  \mathrm{ReLU}\!\left(
  W^{(k)} \sum_{u \in \mathcal{N}(v)} h_u^{(k-1)} + b^{(k)}
  \right)
  \right].
\end{equation}

Each pathway produces layer-wise MLP scores, denoted by $s_k^{(1)}$ and
$s_k^{(2)}$. The final predicted centrality score for node $i$ is computed as
\begin{equation}
  \hat{y}_i =
  \left(\sum_{k=1}^{L} s_k^{(1)}\right)
  \odot
  \left(\sum_{k=1}^{L} s_k^{(2)}\right),
\end{equation}
where $\odot$ denotes element-wise multiplication.
The element-wise product is used as a directional gating mechanism between the two pathways. Betweenness centrality depends on whether a node can simultaneously receive and forward shortest-path flow. A high score should therefore require agreement between incoming and outgoing structural representations. In contrast, summation can assign a high score even when only one pathway is active. The product encourages agreement between the two directional views and is evaluated against alternative fusion choices in the ablation study reported in Table~\ref{tab:ablation}.
\subsubsection{Closeness Centrality GNN}

The closeness model uses a single pathway with $L=7$ layers. Its final node score is obtained by summing the layer-wise MLP scores:
\begin{equation}
  \hat{y}_i = \sum_{k=1}^{7} s_k .
\end{equation}

The input includes a degree-normalized adjacency matrix
$A_{\text{mod}}=D^{-1}A$, which provides scale-normalized local connectivity and avoids zero feature vectors during training. Closeness is evaluated on undirected graphs to avoid degenerate rankings caused by unreachable nodes in sparse directed graphs.

\subsection{Hyperparameter Selection}
\label{sec:hyperparams}

Hyperparameters are selected by grid search on the validation set using mean Kendall $\tau$ as the selection criterion. The search includes learning rate, dropout, hidden size, and weight decay. The best configuration is then retrained and evaluated on the held-out test set.

\begin{table}[H]
\centering
\begin{tabular}{lcc}
\toprule
\textbf{Parameter} & \textbf{Betweenness} & \textbf{Closeness} \\
\midrule
Learning rate   & $10^{-3}$  & $10^{-3}$ \\
Dropout         & $0.4$      & $0.2$ \\
Hidden size     & $40$       & $40$ \\
Weight decay    & $0.01$     & $0.00$ \\
Epochs          & 100        & 50 \\
Batch size      & 16         & 16 \\
Train graphs    & 1{,}400    & 3{,}500 \\
Val graphs      & 200        & 500 \\
Test graphs     & 400        & 1{,}000 \\
\bottomrule
\end{tabular}
\caption{Final training hyperparameters selected by validation performance.}
\label{tab:hyperparams}
\end{table}

The best validation performance is obtained with learning rate $10^{-3}$ and hidden size $40$ for both models. Betweenness uses stronger regularization with dropout $0.4$ and weight decay $0.01$, whereas closeness performs best with dropout $0.2$ and no weight decay.

\subsection{Loss Function}
Both models are trained using a pairwise ranking loss rather than mean squared error, because the learning objective is to preserve the centrality-induced ordering of nodes. For each graph, $M=N\times20$ node pairs are sampled. The loss is

\begin{equation}
  \mathcal{L}
  =
  \frac{1}{M}
  \sum_{m=1}^{M}
  \max\!\left(
  0,\;
  -r_m(\hat{y}_{i_m}-\hat{y}_{j_m})
  + \mathrm{margin}
  \right),
\end{equation}

where $r_m \in \{-1,+1\}$ indicates the true ordering between nodes $i_m$ and $j_m$. We use \texttt{MarginRankingLoss} with a margin equal to $1.0$.

\section{Experimental Results}
\label{sec:results}

\subsection{Training Convergence and Rank Preservation}

The quality of the learned node rankings is evaluated using Kendall's $\tau$ rank correlation~\cite{kendall1938}. 
We use Kendall's $\tau$ rather than mean squared error because the objective is to preserve the relative ordering of nodes according to centrality, rather than to predict exact numerical centrality values.

\begin{figure}[H]
 \centering
 \includegraphics[width=\linewidth]{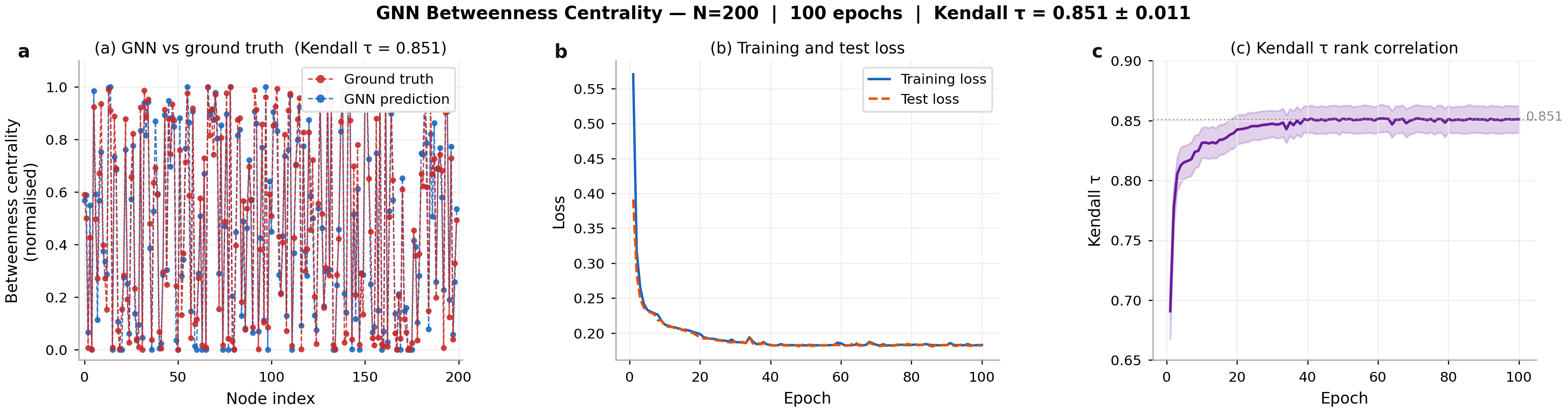}
 \caption{Betweenness centrality approximation on unseen Erd\H{o}s--R\'enyi test graphs. 
 (a)  Comparison between exact NetworkX betweenness centrality and GNN predictions on a representative test graph, with Kendall $\tau = 0.852$. 
 (b) Training and test loss over 100 epochs, showing stable convergence and a small generalization gap. 
 (c) Kendall rank correlation during training, reaching a final test performance of $\tau = 0.851 \pm 0.011$.}
 \label{fig:bet_training}
\end{figure}

Figure~\ref{fig:bet_training} shows the training behaviour of the betweenness model. The training and test losses decrease steadily over 100 epochs and remain close throughout training, with a final relative difference of $0.006$. This indicates stable convergence without clear signs of overfitting. On a representative unseen test graph, the betweenness model achieves Kendall $\tau = 0.852$.
The mean performance is $\tau = 0.851 \pm 0.011$ on unseen Erd\H{o}s--R\'enyi graphs.

\begin{figure}[H]
  \centering
  \includegraphics[width=\linewidth]{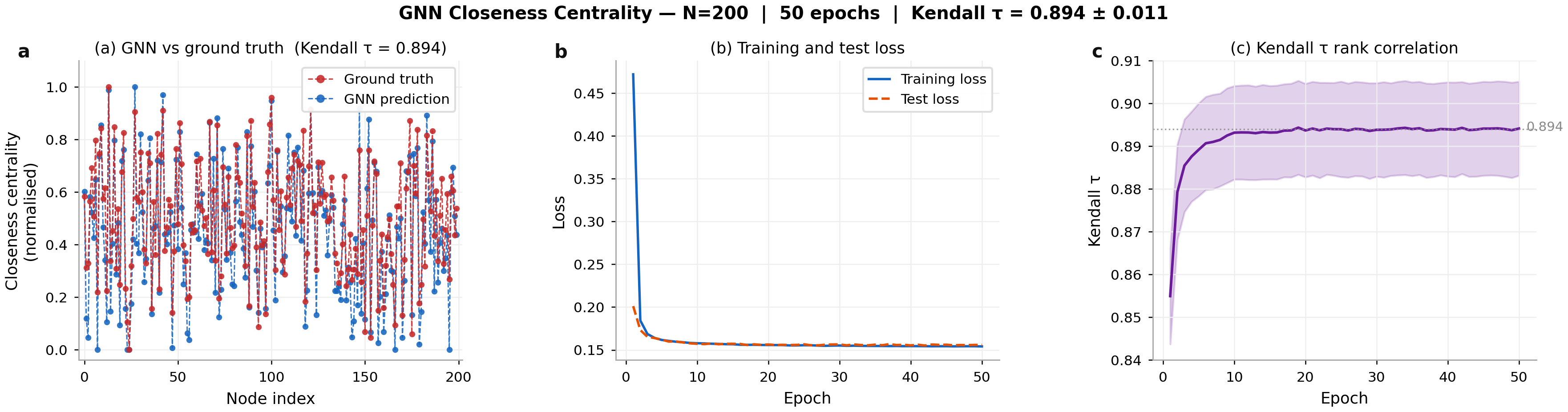}
  \caption{Closeness centrality approximation on unseen Erd\H{o}s--R\'enyi test graphs. 
  (a) Comparison between exact NetworkX closeness centrality and GNN predictions on a representative test graph. 
  (b) Training and test loss over 50 epochs, showing rapid convergence and nearly identical train--test behaviour. 
  (c) Kendall rank correlation during training, reaching a final test performance of $\tau = 0.894 \pm 0.011$.}
 \label{fig:close_training}
\end{figure}

Figure~\ref{fig:close_training} shows the corresponding results for closeness centrality. The closeness model converges rapidly within the first few epochs, and the training and test losses remain nearly identical at convergence, with a final relative difference of $-0.012$. The model achieves $\tau = 0.894 \pm 0.011$, confirming that the learned ranking is highly consistent with exact closeness centrality on unseen test graphs.

Together, these results show that both models preserve node rankings effectively while maintaining small train--test gaps. The comparison with exact NetworkX centrality values further confirms that the models capture the full ranking structure, not only the most central nodes.

\subsection{Ablation Study}
\label{sec:ablation}

To assess the contribution of the main architectural choices, we perform a compact ablation study for betweenness centrality. The variants are trained under the same reduced setting using $N=200$ Erd\H{o}s--R\'enyi graphs and are evaluated on both ER and GRP test graphs.

\begin{table}[H]
\centering
\begin{tabular}{lcc}
\toprule
\textbf{Variant} & \textbf{ER $\tau$} & \textbf{GRP $\tau$} \\
\midrule
\multicolumn{3}{l}{\textit{Architecture and fusion ablation, $L=5$}} \\
Single pathway, $L=5$ &
$0.471 \pm 0.031$ &
$0.390 \pm 0.046$ \\
Dual pathway + sum, $L=5$ &
$0.827 \pm 0.012$ &
$\mathbf{0.679 \pm 0.025}$ \\
Dual pathway + product, $L=5$ &
$\mathbf{0.846 \pm 0.012}$ &
$0.493 \pm 0.039$ \\
\midrule
\multicolumn{3}{l}{\textit{Depth ablation using dual-pathway sum fusion}} \\
Dual pathway + sum, $L=3$ &
$0.781 \pm 0.018$ &
$0.477 \pm 0.033$ \\
Dual pathway + sum, $L=5$ &
$\mathbf{0.798 \pm 0.014}$ &
$0.351 \pm 0.040$ \\
Dual pathway + sum, $L=7$ &
$0.722 \pm 0.020$ &
$\mathbf{0.488 \pm 0.034}$ \\
\bottomrule
\end{tabular}
\caption{Ablation study for the betweenness GNN. The first block compares pathway and fusion choices at fixed depth $L=5$, while the second block evaluates message-passing depth using dual-pathway sum fusion.}
\label{tab:ablation}
\end{table}

The ablation results show that the dual-pathway design is essential: The single-pathway model performs substantially worse on both ER and GRP graphs. Product fusion gives the best in-distribution ER performance, while sum fusion transfers better to GRP graphs. The depth ablation shows that increasing depth does not monotonically improve performance, suggesting that message-passing depth interacts with topology shift.

\subsection{Baseline Comparison}
\label{sec:baselines}

We compare the proposed GNN models against random ranking, degree centrality, and sampling-based Brandes approximation. Random ranking gives $\tau \approx 0$, confirming that the evaluation metric behaves as expected. Degree centrality is a strong inexpensive baseline, while Brandes sampling provides a standard approximation baseline for betweenness centrality. Since Brandes sampling is defined for betweenness approximation, it is not reported for closeness centrality.

Table~\ref{tab:baseline_comparison} reports the baseline comparison on ER graphs. Degree centrality performs strongly on ER graphs, achieving $\tau = 0.886$ for betweenness and $\tau = 0.923$ for closeness. This is expected because ER graphs are relatively homogeneous, and degree is already highly correlated with shortest-path-based centrality. The GNN models achieve comparable ranking accuracy, with $\tau = 0.851 \pm 0.011$ for betweenness and $\tau = 0.894 \pm 0.011$ for closeness, while avoiding explicit shortest-path computation at inference time.

\begin{table}[H]
\centering
\begin{tabular}{lccc}
\toprule
\textbf{Method} &
\textbf{Betweenness $\tau$} &
\textbf{Closeness $\tau$} &
\textbf{Time (ms)} \\
\midrule
Random Ranking &
$-0.001 \pm 0.050$ & $-0.002 \pm 0.052$ & 0 \\
Degree Centrality &
$0.886 \pm 0.010$ & $0.923 \pm 0.010$ & 0.07 \\
Brandes $k=10$ &
$0.286 \pm 0.044$ & --- & 7.9 \\
Brandes $k=20$ &
$0.378 \pm 0.038$ & --- & 15.5 \\
Brandes $k=50$ &
$0.552 \pm 0.032$ & --- & 35.2 \\
\textbf{GNN (ours)} &
$\mathbf{0.851 \pm 0.011}$ &
$\mathbf{0.894 \pm 0.011}$ &
23.8 / 9.7 \\
\bottomrule
\end{tabular}
\caption{Baseline comparison on Erd\H{o}s--R\'enyi test graphs with $N=200$. Brandes sampling is reported only for betweenness centrality.}
\label{tab:baseline_comparison}
\end{table}

To address whether the Brandes baseline behaves similarly under topology shift, we also evaluate sampling-based betweenness approximation on BA and GRP graphs. Table~\ref{tab:brandes_topology} shows that Brandes performance improves with the number of sampled pivots on all graph families, but remains below the mixed-trained GNN on BA and GRP graphs. At $k=50$, Brandes reaches $\tau = 0.747$ on BA and $\tau = 0.629$ on GRP, whereas the mixed-trained GNN achieves $\tau = 0.920$ and $\tau = 0.861$, respectively.

\begin{table}[H]
\centering
\begin{tabular}{lccc}
\toprule
\textbf{Method} & \textbf{ER $\tau$} & \textbf{BA $\tau$} & \textbf{GRP $\tau$} \\
\midrule
Random Ranking &
$0.003 \pm 0.050$ &
$0.002 \pm 0.048$ &
$0.002 \pm 0.046$ \\
Degree Centrality &
$0.869 \pm 0.012$ &
$0.865 \pm 0.011$ &
$0.868 \pm 0.015$ \\
Brandes $k=10$ &
$0.201 \pm 0.043$ &
$0.514 \pm 0.031$ &
$0.380 \pm 0.037$ \\
Brandes $k=20$ &
$0.274 \pm 0.044$ &
$0.606 \pm 0.023$ &
$0.456 \pm 0.031$ \\
Brandes $k=50$ &
$0.434 \pm 0.036$ &
$0.747 \pm 0.017$ &
$0.629 \pm 0.026$ \\
Mixed GNN (ours) &
$\mathbf{0.878}$ &
$\mathbf{0.920}$ &
$\mathbf{0.861}$ \\
\bottomrule
\end{tabular}
\caption{Betweenness baseline comparison across graph families. Brandes sampling improves with larger sampling budgets, but the mixed-trained GNN achieves stronger ranking accuracy on all three graph families.}
\label{tab:brandes_topology}
\end{table}

These results show that degree centrality is a competitive baseline on homogeneous and synthetic graph families, but it is not a learned approximation method and does not provide a trainable representation for transfer. Brandes sampling improves as the number of pivots increases, especially on BA graphs, but still remains below the mixed-trained GNN. Overall, the results support the use of structurally diverse training for learning transferable betweenness representations, while closeness centrality remains more sensitive to topology shift.

\subsection{Cross-Distribution Generalization}
\label{sec:generalization}

A central question of this study is whether GNN-based learning systems acquire transferable representations of node importance or simply adapt to the graph family used during training. To examine this, we evaluate models trained on Erd\H{o}s--R\'enyi (ER) graphs on structurally different Barab\'asi--Albert (BA) and Gaussian Random Partition (GRP) networks.

\begin{figure}[H]
\centering
\includegraphics[width=0.88\textwidth]{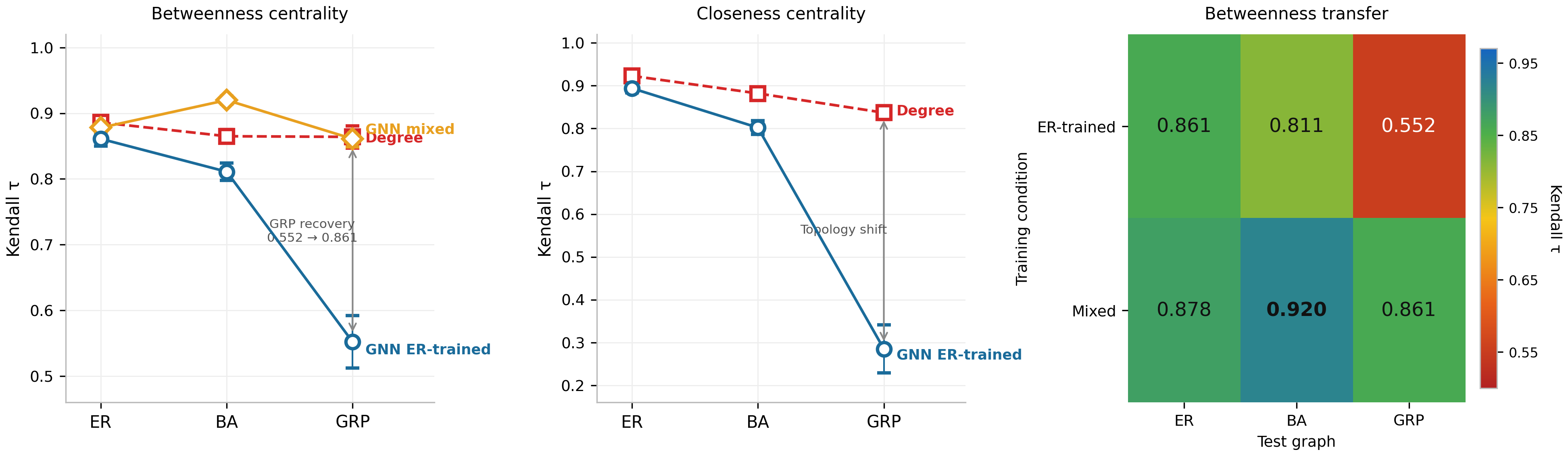}
\caption{Cross-distribution generalization across graph topologies.
(a) Betweenness centrality transfer across ER, BA, and GRP graphs.
(b) Closeness centrality is more sensitive to topology shift, especially on GRP graphs.
(c) Heatmap summary showing that mixed-distribution training improves betweenness transfer.}
\label{fig:generalization_heatmap}
\end{figure}

As shown in Figure~\ref{fig:generalization_heatmap}, the ER-trained betweenness model transfers reasonably well to BA graphs, with Kendall correlation decreasing from $\tau = 0.861$ on ER graphs to $\tau = 0.811$ on BA graphs. However, performance drops substantially on GRP graphs, reaching only $\tau = 0.552$. This indicates a strong topology-shift effect, particularly for community-structured networks where inter-community bridge nodes play an important role in shortest-path routing.

Mixed-distribution training substantially improves this behavior. A single model trained on ER, BA, and GRP graphs achieves $\tau = 0.878$ on ER, $\tau = 0.920$ on BA, and $\tau = 0.861$ on GRP networks without retraining. The largest improvement occurs on GRP graphs, where performance increases from $\tau = 0.552$ to $\tau = 0.861$. This suggests that exposure to diverse graph structures helps the model learn more general centrality patterns rather than topology-specific shortcuts.

The closeness model shows stronger sensitivity to topology shift. While the ER-trained closeness model performs well on ER graphs ($\tau = 0.880$) and reasonably on BA graphs ($\tau = 0.802$), it drops sharply on GRP graphs ($\tau = 0.285$). This indicates that closeness centrality is more affected by community-level reachability patterns and may require greater structural diversity during training to achieve robust transfer. 
The stronger degradation of closeness centrality can be explained by its dependence on global distance structure. In ER graphs, shortest-path distances are relatively homogeneous, and local connectivity often provides a useful proxy for closeness. In contrast, GRP and power-grid networks exhibit modular or spatial structure, larger effective diameters, and more heterogeneous distance distributions. Nodes with similar local neighborhoods may therefore have very different average distances to the rest of the graph depending on their position within or between communities. This makes closeness more sensitive to topology shift than betweenness in the present experiments.

\subsection{Transfer to Real-World Networks}
\label{sec:real_world}

\begin{figure}[H]
\centering
\includegraphics[width=0.78\linewidth]{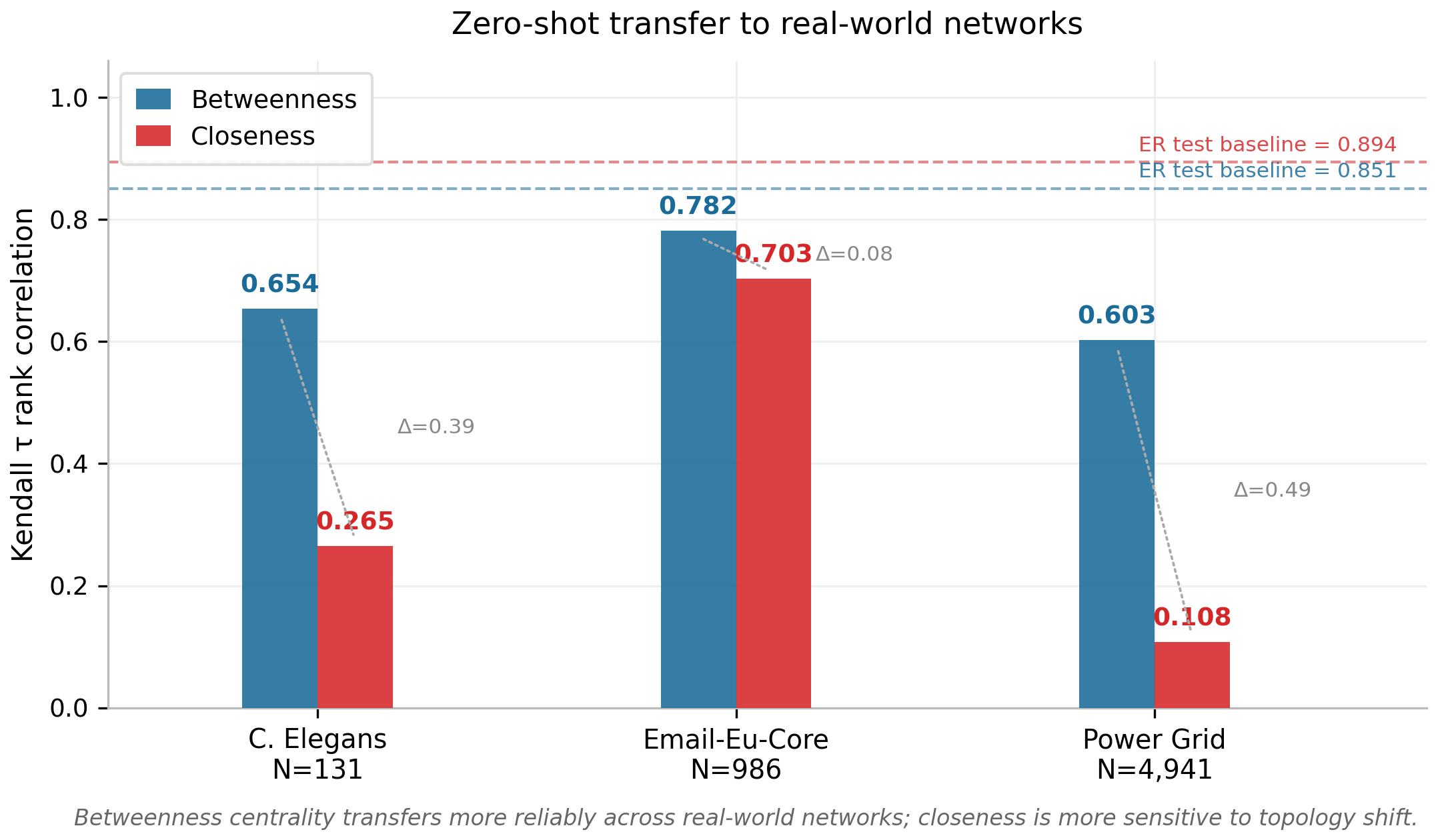}
\caption{Zero-shot transfer to real-world networks. Models trained on synthetic graphs were evaluated directly on three real-world networks without retraining. Betweenness centrality transfers more reliably across datasets, whereas closeness centrality is substantially more sensitive to topology shift.}
\label{fig:real_world_validation}
\end{figure}

To evaluate whether the learned GNN representations transfer beyond synthetic graph families, the trained models were tested on three real-world networks without retraining: the \textit{C. Elegans} neural network, Email-Eu-Core, and the Western United States power grid. As shown in Figure~\ref{fig:real_world_validation}, betweenness centrality transfers moderately well across all three networks, with Kendall correlations ranging from $\tau = 0.603$ to $\tau = 0.782$. In contrast, closeness centrality is much more sensitive to topology shift, with performance ranging from $\tau = 0.108$ to $\tau = 0.703$.
These results indicate that GNNs trained on synthetic graphs can learn reasonably transferable representations for betweenness-based node ranking, whereas closeness-based ranking remains more difficult to generalize to real-world networks. This pattern is consistent with the synthetic generalization results in Section~\ref{sec:generalization}, where closeness centrality also showed substantially greater sensitivity to changes in graph topology.
The especially low closeness performance on the power-grid network is consistent with this interpretation, since power-grid graphs are spatially embedded and typically contain long paths and sparse connectivity.

\subsection{Scalability and Computational Efficiency}
We evaluate computational efficiency by comparing the inference time of the learned GNN approximator with exact NetworkX betweenness computation. The reported speedups refer only to inference time; training is performed once and its cost is amortized when the trained model is applied to many graphs.

As shown in Table~\ref{tab:scalability}, the advantage of GNN inference increases with graph size. At $N=200$, the GNN is $14\times$ faster than exact computation, while at $N=1000$ the speedup reaches $97.7\times$ (149.5\,ms versus 14605.4\,ms). Importantly, the model also preserves high ranking accuracy at this scale, achieving Kendall $\tau = 0.821 \pm 0.004$ on the $N=1000$ test graphs.

\begin{table}[H]
\centering
\begin{tabular}{rcccc}
\toprule
$N$ & \textbf{NetworkX (ms)} & \textbf{GNN (ms)} &
\textbf{Speedup} & \textbf{Kendall $\tau$} \\
\midrule
50   & $5.2 \pm 0.9$       & $2.6 \pm 0.4$     & $2\times$  & $0.845 \pm 0.029$ \\
100  & $30.5 \pm 9.4$      & $3.7 \pm 0.4$     & $8\times$  & $0.844 \pm 0.021$ \\
200  & $139.3 \pm 5.0$     & $9.8 \pm 1.8$     & $14\times$ & $0.851 \pm 0.011$ \\
500  & $1828.7 \pm 58.2$   & $27.2 \pm 4.5$    & $67\times$ & $0.712 \pm 0.012$ \\
1000 & $14605.4 \pm 431.7$ & $149.5 \pm 7.7$
& $\mathbf{97.7\times}$ & $0.821 \pm 0.004$ \\ 
\bottomrule
\end{tabular}
\caption{Scalability comparison between exact NetworkX betweenness computation and GNN inference. All times are mean $\pm$ standard deviation. Kendall $\tau$ at $N=1000$ is evaluated on a reduced test set due to the higher cost of exact ground-truth computation.}
\label{tab:scalability}
\end{table}

For $N=1000$, we repeated the timing experiment after model warm-up and report stabilized inference times, which substantially reduces the variance observed in preliminary measurements.

\subsection{Training at $N=5{,}000$}
\label{sec:N5000}
To assess scalability beyond the primary $N=200$ setting, we trained a dedicated betweenness GNN on Erd\H{o}s--R\'enyi graphs with $N=5{,}000$ nodes and edge probability $p=0.001$. Exact betweenness labels were generated for 300 graphs using an NVIDIA A100 GPU. Label generation required 6.65 hours in total, while model training completed in 8 minutes over 50 epochs.

The dataset was split into 210 training, 30 validation, and 60 test graphs. The same architecture and hyperparameters selected at $N=200$ were used without modification. The best model achieved a validation Kendall correlation of $\tau = 0.9385 \pm 0.0012$ and a test correlation of $\tau = 0.9380 \pm 0.0011$. These results show that the proposed GNN architecture remains effective at substantially larger graph sizes and that the learned approximation framework scales beyond the primary $N=200$ setting.

\section{Conclusion}
\label{sec:Conclusion}
This study investigated Graph Neural Networks as scalable neural approximators for centrality-based node ranking across different network topologies. The proposed models achieve mean Kendall correlations of $\tau = 0.851 \pm 0.011$ for betweenness and $\tau = 0.894 \pm 0.011$ for closeness on unseen Erd\H{o}s--R\'enyi graphs. A dedicated large-scale betweenness model trained on graphs with $N=5{,}000$ nodes further achieves $\tau = 0.938$, while GNN inference provides up to a $97.7\times$ speedup over exact computation.
The main finding is that training diversity improves transferability. Models trained only on Erd\H{o}s--R\'enyi graphs degrade under topology shift, especially on community-structured networks. In contrast, mixed-distribution training on Erd\H{o}s--R\'enyi, Barab\'asi--Albert, and Gaussian Random Partition graphs produces more robust betweenness representations across graph families. This suggests that structural diversity acts as an implicit regularizer, reducing reliance on topology-specific shortcuts.
Zero-shot validation on real-world networks further shows that synthetic training can yield transferable representations for betweenness centrality, although closeness centrality remains more sensitive to topology shift. Overall, the results support GNN-based centrality approximation as a scalable learning-based approach for node ranking in large-scale graph-structured data.

\section{Limitations and Future Work}

Several limitations of the proposed learning-based approximation framework should be noted. First, closeness centrality is more sensitive to topology shift than betweenness centrality, with substantial degradation on community-structured and some real-world networks. This suggests that closeness may require greater structural diversity during training or topology-aware model design. Second, although synthetic graph families allow controlled experimentation and exact supervision, they do not fully capture the complexity of real-world networks. Finally, scalability was evaluated up to $N=5{,}000$ nodes; experiments on larger and more heterogeneous networks would provide a more complete assessment of computational performance.
Future work should therefore focus on improving transfer from synthetic to real-world networks. Promising directions include mixed-distribution training for closeness centrality, incorporating real network data during training, domain adaptation, and topology-aware pretraining. More broadly, graph foundation models and pretraining strategies may offer useful directions for learning transferable centrality representations across heterogeneous graph domains.

\section*{Generative AI Use Disclosure}
During the preparation of this manuscript, the authors used AI-assisted tools, including Grammarly, for grammatical proofreading and sentence-level language refinement, and Claude AI by Anthropic for Latex formatting assistance and figure-code organization. After using these tools, the authors reviewed, edited, and validated all content, and take full responsibility for the final manuscript.
\section*{Data Availability}
All code, training scripts, and experiment pipelines used in this study are publicly available at:
\url{https://github.com/Samra771/graph_ranking-samra_dev}.

\section*{Declaration of Competing Interest}
The authors declare that they have no known competing financial interests or personal relationships that could have appeared to influence the work reported in this paper.




\bibliographystyle{elsarticle-num}
\bibliography{references}
\end{document}